\documentclass[11pt,a4paper]{article}
\usepackage{authblk}
\usepackage[hyperref]{acl2017}
\usepackage{times}
\usepackage{latexsym}

\usepackage{url}
\aclfinalcopy 

\title{Generating Challenge Datasets for Task-Oriented Conversational Agents through Self-Play}

\author[1]{Sourabh Majumdar}
\author[2]{Serra Sinem Tekiro\u{g}lu}
\author[2]{Marco Guerini}

\affil[1]{The State University of New York, NY 14260, Buffalo, USA \protect\\
\texttt{smajumda@buffalo.edu}}
\affil[2]{Fondazione Bruno Kessler, Via Sommarive 18, Povo, Trento, Italy
\protect\\ \texttt{tekiroglu@fbk.eu, guerini@fbk.eu}}

\date{}
\begin{document}
\maketitle
\begin{abstract}

End-to-end neural approaches are becoming increasingly common in conversational scenarios due to their promising performances when provided with sufficient amount of data. In this paper, we present a novel methodology to address the interpretability of neural approaches in such scenarios by creating challenge datasets using dialogue self-play over multiple tasks/intents. Dialogue self-play allows generating large amount of synthetic data; by taking advantage of the complete control over the generation process, we show how neural approaches can be evaluated in terms of unseen dialogue patterns. We propose several out-of-pattern test cases each of which introduces a natural and unexpected user utterance phenomenon. As a proof of concept, we built a single and a multiple memory network, and show that these two architectures have diverse performances depending on the peculiar dialogue patterns.

\end{abstract}

\section{Introduction}


In recent years, there has been an increasing research on neural approaches for conversational systems. 
Such approaches 
include the realization of full end-to-end systems \cite{serban2016building,bordes2016learning}, the capacity to incorporate or query structured knowledge sources into neural architectures \cite{eric2017key}, the use of zero-shot learning or of synthetic dialogues generation techniques to mitigate effort of domain portability \cite{zhao2018zero,guerini2018toward}. 
Although state-of-the-art neural models achieve high performances in many domains, the sheer size of data they require represents a bottleneck, especially for under-resourced dialogue domains. In addition, it is hard to interpret their behaviour since neural models are intrinsically opaque. In this paper we propose a novel methodology to address the aforementioned problems by synthetically creating conversation datasets with peculiar characteristics. 
In particular, we focus on (i) scalable and portable approaches for generating dialogues from scratch involving complex and structured knowledge, and 
(ii) strategies for isolating and evaluating specific reasoning capabilities of neural architectures using synthetic \textit{challenge sets}. To this end, we utilize dialogue self-play strategies~\cite{shah2018building} simulating task oriented dialogues between a conversational agent and a user. 
Dialogue simulation grants complete control over the generated data allowing building special test cases, each of which introduces a natural and unexpected user utterance phenomenon, that has never been seen at training time. It should be noted that the use of natural data would require \textit{manual} annotation/selection of examples for each phenomenon of interest, making this goal intractable.

Another problem with current datasets is the lack of exhaustiveness: they usually focus on one specific domain intent. There are few datasets covering multiple intents in the same scenario, but usually these intents are very different from one another -- e.g. weather forecast and play music in a car scenario~\cite{eric2017key}. For this reason, we choose a compelling low resource dialogue domain, i.e. Banking, that allows us to bring together multiple tasks (e.g. transfer money, check balance, block card). 

\section{Related Work}\label{sec:related-work}
Our focus is on synthetic data generation techniques for low-resource domains and for investigating the learning capabilities of neural conversational agents. For this reason, we will discuss some artificial conversational datasets, testing strategies for dialogue models, and the main dialogue systems approaches.

\noindent\textbf{Artificial Datasets.} Many conversational datasets rely on crowd sourcing~\cite{jurvcivcek2011real,kelley1984iterative}, cooperating corporations~\cite{raux2005let}, already available records~\cite{lowe2015ubuntu,danescu2011chameleons}, or participation with the real world~\cite{yu2015incremental,zue1994pegasus}. 
Moreover, available task oriented dialogue datasets are mainly focused on very specific domains, such as restaurant reservation ~\cite{jurvcivcek2011real,henderson2014second}, flight booking~\cite{zue1994pegasus}, bus information systems~\cite{raux2005let} and giving directions to rooms in a hotel~\cite{yu2015incremental}. Since the collection of whole dialogues is usually very expensive and time consuming, there have been efforts in building methodologies that allow fast and cheap data collection \cite{shah2018building,kelley1984iterative}.
Artificial data generation is an effective methodology for obtaining well defined training datasets. 
\citet{bordes2016learning} uses a simulator for this purpose. Other approaches artificially outline the conversational flow of their examples and then use crowdsourcing to convert dialogue turns into natural language~\cite{shah2018building}. 

\noindent\textbf{Testing Neural Dialog Models.}
With the rise of Neural NLP, interpretability has become a major issue. 
\citet{belinkov:2018:tacl} survey various methods addressing interpretability. In particular, one line of research deals with challenge sets \cite{lehmann1996tsnlp}. While the majority of NLP datasets reflects a natural distribution of language phenomena, challenge sets are meant to restrict their focus on a specific phenomenon (quantifiers, plurals, anaphora, etc.) at a time \cite{cooper1996using}. Analyzing the ability to deal with a specific phenomenon allows evaluating the systems in a more principled way. Challenge datasets have been applied to diverse tasks, such as Natural Language Inference \cite{wang2018glue} and Machine Translation \cite{king1990using}. 
These datasets offer insight if a model is capable of handling a specific line of reasoning from training data to specific structures at test time. 

\noindent\textbf{Methods for conversational scenarios.} Many methods have been proposed to deal with conversational scenarios. 
\textit{Rule Based Systems} are the simplest to implement for the task oriented setting when the flow of the dialogue is already known. 
They tend to be highly brittle to patterns not seen during their construction or to porting to new domains \cite{marietto2013artificial}. 
\textit{Information Retrieval Methods} usually imply the two main approaches, namely TF-IDF and Nearest Neighbor models~\cite{isbell2000cobot,jafarpour2010filter,ritter2011data,sordoni2015neural}. More recently, \textit{Neural Network Models} have been heavily employed for conversational agents. Sequence-to-sequence models~\cite{vinyals2015neural} 
perform well for short conversations but fail in longer ones~\cite{hochreiter1997long,cho2014properties}. 
Hierarchical Encoder Decoder Models~\cite{serban2016building} are an extension of the sequence-to-sequence models. 
They handle the context in a separate RNN and use this context for response generation. 
Finally, Latent Variable Hierarchical Encoder Decoder models~\cite{serban2017hierarchical} represent a further improvement of the Hierarchical Encoder Decoder Model. 
Other recent approaches have focused on Memory Networks, that use an external memory component build into the system to store long term contexts \cite{Weston2014MemoryN}. In particular, end-to-end Memory Networks \cite{sukhbaatar2015end}, an extension where every component is trained in an end-to-end fashion, showed promising results in task oriented dialogues \cite{bordes2016learning}.

\begin{table*}
 \centering
 \begin{tabular}{l|l} 
 \hline 
\textbf{ Logical Form} & \textbf{Example Annotation} \\
 \hline 
 \textit{BOT: How can I help you today ?} & \textit{BOT: How can I help you today ?}  \\
 inform\_intent = transfer & I need to send some money. \\
 inform\_intent = transfer \{amount\} & I want to transfer \{amount\}  \\
 inform\_intent = transfer \{partner\} &  Can I send money to \{partner\}? \\
 inform\_intent = transfer \{partner\} \{amount\} & I would like to send \{amount\} to \{partner\}. \\
 \hline
  \textit{BOT: Who is the recipient?} & \textit{BOT: Who is the recipient?} \\
 inform \{partner\} & It is \{partner\}. \\	
 \hline
  \textit{BOT: What is the transfer amount?} & \textit{BOT: What is the transfer amount?} \\
 inform \{amount\} & Roughly \{amount\}. \\	
 \hline
 \end{tabular}
 \caption{Automatically generated logical forms provided to annotators and annotated template samples for a Money Transfer intent with 2 slots. Bot request is provided to annotators to give better context.}
 \label{TAB:template_ex}
\end{table*}

 \section{Data Generation through Self-Play}\label{sec:Data-Generation}

Unlike the natural challenge sets discussed in Section~\ref{sec:related-work}, our focus is on testing structured reasoning in conversational context, by using synthetic dialogue generation to grant that phenomena under investigation are present only at test time and not at training time. To our knowledge, this is the first attempt to build challenge datasets in an artificial way and to use them for the dialogue domain. Natural data would not be suitable for our purpose since the challenging test phenomena can also be found in the training set. 
In particular, our dataset is constructed using a dialogue self-play approach~\cite{shah2018building}. This approach suggests that we can always simulate a conversation if we treat it like a game. During the simulations, we instantiate two bots;
the system and the user. For each conversation, we pick up a user profile, then user and system bots carry on the conversation through pseudo-language \textit{\textbf{actions}} regarding the user profile.



Therefore, every conversation is represented as an exchange of actions between the two agents. Each action contains the information on who performed it and what values the agent provided. 
For each dialog for a chosen intent, the respective system bot asks the relevant slots and the user bot provides the appropriate slot values. The system bot then checks the values provided and issues the next requests accordingly. Both API calls and slot requests are performed through actions.
To convert these actions into actual dialogues, we conducted an annotation task with 5 annotators. First, we converted each possible action for all the intents into logical forms (uninstantiated pseudo-code) and asked the annotators to provide natural language templates for each logical form without changing the entity placeholders. 
We then used the language templates 
to create natural language utterances by replacing the placeholders with the actual values. In Table~\ref{TAB:template_ex}, we give a few examples of the logical representations provided to annotators and the template they wrote. The conversion to the final plain text has been done by filling annotated templates with a small KB (the user profile). 
This approach is different from the one proposed by~\citet{shah2018building} that uses instantiated pseudo-code since the beginning: it requires much more data annotation and makes it more difficult to detach surface realization from dialogue structure for building challenge sets.

The advantages of our synthetic data generation is twofold. First, it helps us to achieve a better coverage since reducing dialogues to an exchange of actions allows having a formal and simple visualization of all the generated dialogue flows - this is not the case with WoZ data collections \cite{kelley1984iterative}. Second, by instantiating each dialogue with different natural language templates, we can create dialogues that have the same structure but different wording.

\subsection{The System and User Bot Design}



We designed the user and the system bots using a finite state machine approach~\cite{hopcroft2001introduction}, assuming that each dialogue is a flow between states and each utterance is a move from one state to the next. For our experiments, each of these states handles one particular slot. 
Whenever a user bot provides some information through its action, the system bot changes the state accordingly and performs its own action in response. The user bot then deals with the system action, performs its own and the cycle continues. The dialogue concludes when the system bot reaches an end state and issues an end call action. 

\subsection{Banking Domain Intents}
The user intents/tasks that the dialogues are built upon are in the Banking Domain. We selected several intents to create a dataset that contains: 
i) conversations that are varied and diverse, and 
ii) conversations that use similar sentences and slots but to complete different tasks.

Each dialogue is initiated by the system bot asking the first question. To add variability 
we randomized the number of slots provided by the user in its response, similar to verbosity user trait in~\cite{shah2018building}. After these initial turns the system bot ask the respective missing slots. Apart from asking the missing slot, the system bot also validates the value of the slots by calling respective APIs, since by design the User Bot may provide incorrect values and the System Bot is designed to handle all these possible cases. In Table \ref{TAB:block_card_ex}, we give an example dialogue 
to demonstrate how API intensive these dialogues might become. %
Note that, for each interaction we have a specific user profile, i.e. the possible entities for slots are predetermined. For example, ``partner list" for money transfer is fixed, and a user cannot provide a random name. 
Additionally, dialogues are formed with respect to policy restrictions such that i) the slots for each intent are asked in a strict order, ii) each slot value should be validated through an api call, iii) there are maximum 3 consecutive invalid user utterances after which the dialogue is terminated, iv) the final API call for intent execution randomly returns a success or failure. The valid/invalid user utterances are chosen randomly in each turn. We now list the domain intents and their descriptions as follows:




\paragraph{Making a Transfer.} In this intent, \textsc{User} wants to transfer some money. The system is tasked to fill three slots i.e \textit{user account}, \textit{partner name} and \textit{amount}. 

\paragraph{Canceling a Transfer.} This intent handles cases where \textsc{User} wants to cancel a transfer made earlier. The first required slot is the \textit{transaction id} and \textsc{User} may or may not recall it. If the transaction id cannot be filled, \textsc{system} asks for \textit{partner name} and checks for previous transactions. 

\paragraph{Blocking a Card.}
The main slot required for this intent is the \textit{card id}. Similar to the canceling a transfer intent, there are two possible paths for the conversation flow. If \textsc{User} cannot recall the \textit{card id}, \textsc{system} asks another set of slots, which are the \textit{user account} and the \textit{card type}. 

\begin{table}[t!]
 \centering
 \begin{tabular}{|p{\linewidth}|} 
 \hline
\textbf{S} : How can i help you today?\\
\textbf{U} : I am really concerned that i lost the bank card, is its deactivation possible?\\
\textbf{S} : Please enter the card id.\\
\textbf{U} : I am not sure about the card id\\
\textbf{S} : Which account does the card belong to?\\
\textbf{U} : It is associated with my checking account.\\
\textbf{A}: \small{\textsc{ call:check\_account, user\_account:checking}} \\
\textbf{A}: \small{\textsc{response:account\_check, result:failed, message:}}\normalsize`list of available accounts joint , savings'\\
\textbf{S} : I couldn't find the account you are looking for. Your available accounts are joint , savings. Would you like to change your answer?\\
\textbf{U} : Yes\\
\textbf{S} : Which account does the card belong to?\\
\textbf{U} : It must be savings\\
\textbf{A} : \small\textsc{call:check\_account, user\_account:saving}\\
\textbf{A} : \small\textsc{response:account\_check, result:success}\\
\textbf{S} : You have the following cards that belong to your account: prepaid card, credit card. which card would you like to block?\\
\textbf{U} : Can you block the prepaid card?\\
\textbf{A} : \small\textsc{call:check\_card\_name, card\_name:prepaid}\\
\textbf{A} : \small\textsc{response:card\_name\_check, result:success}\\
\textbf{S} : Do yo confirm blocking your card?\\
\textbf{U} : yes\\
\textbf{A} : \small\textsc{call:block\_card, card\_id:savings-prepaid}\\
\textbf{A} : \small\textsc{response:block\_card, result:success}\\
\textbf{S} : Your card has been successfully blocked.\\
\hline
 \end{tabular}
 \caption{A complete Block card example. U corresponds to User, S to System, and A to API calls. }
 \label{TAB:block_card_ex}
\end{table}

\paragraph{Transaction History Query.} This intent is for finding a transaction in \textsc{User}'s transaction history by searching the \textit{note} field. Notes are specific texts giving information about the transaction reason, such as ``Salary" or ``Electricity bill". 

\paragraph{Account Balance Query.} To search for the account balance, \textsc{User} is asked to provide the \textit{user account} name. 

\paragraph{Account Limit Query.} \textsc{User} searches for the maximum transferable amount from his/her account. This intent, together with account balance, serves to evaluate the disambiguation performance of the conversational models when facing semantically similar intents - in fact their dialogue flow and required slots are identical.




\subsection{Challenge Test Set}
We generate a wide variety of test cases using the same dialogue self-play methodology employed for the training data, by modifying the structure of the original dialogues in a principled manner. Each of these test cases represents a particular conversational situation that has never been seen during training but that might happen in real interactions. While we cannot precisely quantify the exhaustiveness of the considered cases in real life scenarios due to the domain restrictions, the synthetic dialogue generation allows us to easily integrate new challenge cases when necessary. Below, we discuss in detail each test case and what function it serves to test. We also present a more conventional test set, i.e. out of templates set, to disclose how much we elevate the difficulty of the task by the challenge test cases. In our experiments, we will disregard Out Of Vocabulary (OOV) cases, i.e. cases where entities of interest at test time are not seen at training time, e.g. the partner name ``Michael", \cite{bordes2016learning}. 

\paragraph{Out of Templates (OOT)} Out of templates test case is constructed by sampling annotation templates into training, development and test sets. By doing so, we can test whether the agent can handle user utterances in an unseen linguistic form. 

\paragraph{Out of Pattern (OOP)} Out of pattern test cases challenge the agents through the conversations with a structure (dialogue flow) that has not been seen during training.
We have constructed five out of pattern cases for the challenge test set. 
Together with each OOP description, we provide an example, where the arrow indicates the answer of the system bot to be predicted, and the part in italic highlights the differences between training and test sets for readability purposes. 

\paragraph{1. Turn Compression}
The training dialogue structure contains a confirmation step after each attempt to fill the current slot with an invalid value. Following a confirming answer from \textsc{User} for changing the slot value, \textsc{System} repeats the slot filling question to \textsc{User}. In the turn compression challenge test case, we concatenate the confirmation and the new slot answers as the user utterance for the change confirmation question of \textsc{System}. The correct system utterance is the validation API call of the new value instead of the slot filling question again. Table \ref{tab:Turn Compression} shows an example where \textsc{System} asks if \textsc{User} wants to change the \textit{partner name}, and during testing the user gives a confirming answer together with the correct \textit{partner name}.

\begin{table}[h!]
\centering
 \begin{tabular}{p{0.95\columnwidth}}
 \hline
 \textbf{Sys} : Partner name is incorrect, would you like to change it?\\
 \textbf{User} : \textit{Yes}\\
 $\rightarrow$\textbf{Sys} : \textit{What is the name of the recipient?}\\
 \hline
 \textbf{Sys} : Partner name is incorrect, would you like to change it?\\
 \textbf{User} : \textit{Yes, change it to Megan}\\
 $\rightarrow$\textbf{Sys} : \textit{API PARTNER CHECK \textbf{Megan}}\\
 \hline
 \end{tabular}
 \caption{Train/test excerpt of Turn Compression}
 \label{tab:Turn Compression}
\end{table}

\paragraph{2. New API}
The new API challenge is designed to evaluate the performance of an agent in terms of its ability of issuing the required API calls with appropriate slot values. To this end, we create intent utterances in training data either without any extra slot value or odd number of slot values while in the new API challenge set, intent utterances contain only even number of slot values. See Table \ref{tab:New API} for an example. 

\begin{table}[h!]
 \centering
 \begin{tabular}{p{0.95\columnwidth}}
 \hline
 \textbf{Sys} : How can i help you today ?\\
 \textbf{User} : I want to see if my salary was credited \\
 $\rightarrow$ \textbf{Sys} : API CHECK SALARY\\
 \hline
 \textbf{Sys} : How can i help you today?\\
 \textbf{User} : I want to see if my salary was credited from Facebook \\
 $\rightarrow$ \textbf{Sys} : API CHECK SALARY \textit{FACEBOOK}.\\
 \hline
 \end{tabular}
 \caption{Train and test excerpt of New API}
 \label{tab:New API}
\end{table}

\paragraph{3. Reordering}
The training dialogues have a particular slot filling order for each intent. In the reordering challenge set, on the other hand, the order of slots in dialogues is shuffled. The purpose of reordering is to evaluate whether an agent is able to generate the system utterance to fill the correct missing slot after all other slots for the current intent have been filled. In the example given in Table \ref{tab:Reordering}, we show the correct system utterance of a new API challenge test case for the missing slot: \textit{user account}.


\begin{table}[h!]
 \centering
 \begin{tabular}{p{0.95\columnwidth}}
 \hline
 \textbf{User} : I'd like to transfer some money.\\
 \textbf{Sys} : From which account?\\
 \textbf{User} : from Savings please.\\
 \textbf{Sys} : Who is the recipient ?\\
 \textbf{User} : It is Michael\\
 $\rightarrow$ \textbf{Sys} : \textit{What is the amount?}\\
 \hline
 \textbf{User} : I'd like to transfer some money.\\	
 \textbf{Sys} : \textit{What is the amount} ? \\ 
 \textbf{User} : 100 euros.\\
 \textbf{Sys} : Who is the recipient ?\\
 \textbf{User} : It is Michael\\
 $\rightarrow$ \textbf{Sys} : \textit{From which account?}\\
 \hline
 \end{tabular}
 \caption{Train and test excerpt of Slots Reordering}
 \label{tab:Reordering}
\end{table}

\paragraph{4. Another Slot} In a natural conversation, \textsc{User} can provide an utterance that is irrelevant to the current turn while including a slot value relevant to another one. 
Example dialogue excerpts for Making a Transfer intent with and without such \textit{another slot} test case are given in Table \ref{tab:Another Slot}.

\begin{table}[h!]
 \centering
 \begin{tabular}{p{0.95\columnwidth}}
 \hline
 \textbf{Sys} : What is the partner name ?\\
 \textbf{User} : \textit{It is Michael.}\\
 $\rightarrow$\textbf{Sys} : \textit{API PARTNER CHECK \textbf{Michael}}\\
 \hline
 \textbf{Sys} : What is the partner name ?\\
 \textbf{User} : \textit{Make the amount to be 10k euros .}\\
 $\rightarrow$\textbf{Sys} : \textit{API AMOUNT CHECK \textbf{10k euros}}\\
 \hline
 \end{tabular}
 \caption{Train/test excerpt for Another Slot case}
 \label{tab:Another Slot}
\end{table}


\paragraph{5. Audit More}
In the training dialogues, after \textsc{System} requests a new slot value for the value changing turn, \textsc{User} is expected to provide an appropriate value for the slot. In the audit more challenge set, \textsc{User} provides other slot values along with the requested slot.
We test if the system can recognize the changed slot value and issue the appropriate API calls for the slots given by \textsc{User}. In the example test dialogue given in Table \ref{tab:Audit More}, \textsc{User} changes the amount along with the partner name slot. The correct response of \textsc{System} would be the API calls for the given partner name and the amount values.

\begin{table}[h!]
 \centering
 \begin{tabular}{p{0.95\columnwidth}}
 \hline
 \textbf{Sys} : What is the partner name ?\\
 \textbf{User} : \textit{Change it to Michael.}\\
 $\rightarrow$\textbf{Sys} : \textit{API PARTNER CHECK \textbf{Michael}}\\
 \hline
 \textbf{Sys} : What is the partner name ?\\
 \textbf{User} : \textit{Change it to Michael and make the amount to be 5k euros.}\\
 $\rightarrow$\textbf{Sys} : \textit{API PARTNER CHECK \textbf{Michael}, AMOUNT CHECK \textbf{10k euros}}\\
 \hline
 \end{tabular}
 \caption{Train and test excerpt of Audit more.}
 \label{tab:Audit More}
\end{table}

We should note that each OOP test can contain only one type of OOP, i.e only turn compression or only new API etc.
However, in one dialog there may be more than one instance of the same OOP.

\section{Neural Models}
For our experiments, we focused on end-to-end Memory Networks \cite{sukhbaatar2015end}, which have been employed in various NLP tasks including non-goal-oriented \cite{dodge2016} and goal-oriented \cite{bordes2016learning} dialogue systems. End-to-end memory networks exploit a memory component to store the dialog history, multiple memory look-ups (hops) as an attention mechanism, and short-term context to predict the next response. They yield promising performances in dialogue tasks outperforming some other end-to-end RNN based architectures. In addition, end-to-end memory networks have been shown to be able to perform non-trivial operations such as issuing API calls to KBs \cite{bordes2016learning}, which are a key element to our scenario. We implemented 2 variations of memory networks in order to test the feasibility of challenge set by analyzing the performance differences of the networks. \\ 


\noindent\textbf{Single End-to-End Memory Network (SMN)} replicates the end-to-end memory network presented by \citet{bordes2016learning} trained on all dialogues from all intents simultaneously. 

\noindent\textbf{Multiple End-to-End Memory Network (MMN)} is an ensemble of 6 different Memory Networks, each trained on a single intent and a $7^{th}$ Memory Network that has the task of selecting the appropriate one for the given dialogue. The training data for the $7^{th}$ memory network is produced by appending a call-memory-network action after the user utterance that informs the intent in a dialog. \\

We implemented MMN and tested against SMN to investigate if sharing information from different intents plays a positive role or creates interference, especially for intents that contain the same slot types and have a high semantic similarity. 

\section{Experiments}

We used a learning rate of 0.001, a batch size of 32, and maximum 20 epochs during the training of both networks. We set the embedding size to be 128 as it has been shown to perform good enough for most NLP problems \cite{bai2009supervised}. For memory networks we empirically used a memory size of 40 
and 3 hops. 






We have generated 200 dialogues per intent and per test case through the methodology explained in Section \ref{sec:Data-Generation}. We first sampled $1/3$ of the templates to linguistically realize the logical form of training dialogues and $1/3$ for development. For the in-template test cases we then randomly sampled turns from training and development to create test dialogues. Instead, for the OOT test configurations we used the remaining $1/3$ of the templates to generate new linguistically unseen test cases. 
Similar to \citet{bordes2016learning}, the networks are evaluated in a ranking, not a generation, task: we test whether for each dialogue turn, the MNs can identify the next (correct) system utterance or API call from a list of candidates. Regarding the evaluation metric, we have used Per-Response Accuracy, which is the number of correct system responses predicted out of all dialogue turns in the whole test set. Each response under test is compared to the predicted response by the model when given a dialogue memory, i.e. context, and a new user utterance. 

Considering the banking domain and task scenarios, for which it is almost impossible to record and collect real human 
interactions, we cannot perform an experiment on real life conversations. Although using WoZ technique and challenge pattern annotation could be applicable, it could not be guaranteed that we can collect reasonable amount of data for the same challenge pattern and we would be obliged to know every pattern in advance. Therefore, we also employ dialogue synthesis for test/challenge set dialogues, which allows for a continuous challenge design by modifying the template combinations and dialogue patterns.

\subsection{Test Cases and Results}

We compare the memory networks against the per-response accuracy 
for In Template (IT henceforth) and Out Of Template setting. In Table~\ref{table_:_OOT}, we show the test results of SMN and MMN models for IT and OOT configurations (including non-OOP and OOP test settings).

\begin{table}[ht!]
 \centering
 \begin{tabular}{p{2.1cm}|rr|rr}
  
 \hline
   & \multicolumn{2}{c|}{IT} &  \multicolumn{2}{c}{OOT}\\
 Test Case & \textit{SMN} & \textit{MMN} & \textit{SMN} & \textit{MMN}\\
  \hline
 Non OOP & 88.62 & \textbf{90.17} & 87.39 & \textbf{88.27}\\
  \hline
 Turn Comp. & 27.80 & \textbf{55.00} & 27.90 & \textbf{54.70}\\
 New API & 7.42 & \textbf{7.83}& \textbf{ 8.17} & 6.67\\
 Reordering & \textbf{54.50} & 45.50& \textbf{54.00} & 41.50\\ 
Another Slot & \textbf{38.00} & 25.00 & \textbf{41.50} & 27.50\\
 Audit More & 15.50 & \textbf{34.00} & 16.00 & \textbf{35.00}\\
 \hline
OOP Avg. & 28.64 & \textbf{33.47} & 29.51 & \textbf{33.07}\\
  \hline

 \end{tabular}
 \caption{In template non-challenge/OOP test results, OOT non-challenge/OOP test results in terms of per-response accuracy.}
 \label{table_:_OOT}
\end{table}



\paragraph{OOP impact.} As expected, OOP cases represent a compelling challenge for our MNs, see Table 
\ref{table_:_OOT}. When we compare the results of the non-OOP and the OOP cases, we observe drastic performance differences, both in IT (88.62 vs. 28.64, 90.17 vs 33.47) and OOT (87.39 vs. 29.51, 88.27 vs. 33.07) settings. Still, in some settings both MNs are able to show reasonable performances and different behaviors on different challenge sets (reordering and another slot for SMN, turn compression and audit more for MMN). 

\paragraph{Single vs Multiple Memory Network.} 
Concerning the IT cases, MMN slightly outperforms SMN in the non-challenge test setting. In addition, it shows a substantial accuracy increase in turn compression and audit more OOP cases. On the other hand, SMN surpasses MMN in reordering and another slot OOP challenges. A similar outlook of performances is observed for the OOT non-challenge and OOP cases aside from new api challenge, which turns out to be the most difficult OOP challenge and will be discussed later. 

We observe that on average MMN outperforms SMN both in IT and OOT cases. One possible explanation is based on how the memory network finds the correct response. The Single Memory Network does so by incorporating all the intents in its selection of the responses. Therefore, it searches for more general responses while the Multiple Memory Network assigns the same task to a specialized Memory network, which is trained on that very specific intent. The specialized memory network is better at finding the correct response since it is trained only on one particular intent data and its search space is implicitly reduced.

As a particular OOP performance comparison, we noticed that SMN is better at selecting the right response during the reordering challenge, which evaluates the ability of the model in learning the necessary slots to accomplish an intent. 

\paragraph{In template vs Out of Template.} 

We found out that there is not a major difference between IT and OOT test case performances (slightly better for SMN and slightly worse for MMN). One possible explanation is that the tests have not been conducted in an OOV setting. Therefore, the SMN and MMN may not learn the linguistic patterns to find entities (templates) but they directly learn to recognize the entities and to predict the API calls accordingly. 

\paragraph{Multiple Out of pattern.} 

As a final experiment, we wanted to inspect the effect of having the challenge phenomenon of interest appearing more than once in a dialogue, such as the example proposed in Table \ref{tab: One vs Multiple OOP}. For this last case we could use only turn compression and audit more, that have a sufficient number of slots to replicate the phenomenon of interest over different slots in the same dialogue. What we observe is that indeed it is difficult for both the SMN and the MMN to handle the case of more than one OOP. 

It can be seen that there is a drop in the performances of both MNs in Table \ref{table : Multple OOP New} as compared to previous challenge tests, only for audit more. This drop can be attributed to the differences of the contexts of the conversations that are present in the memory while selecting the response. Since the context is the previous conversation until the chosen turn, for the first case, i.e. the first example in \ref{tab: One vs Multiple OOP}, the context is a usual sub-dialogue pattern that is seen during the training. 
So the responsibility of the agent is to understand the new unseen user utterance and choose the correct response. However, when we test the second audit-more in the same dialogue, i.e. the second turn in \ref{tab: One vs Multiple OOP} where \textsc{User} changes both the partner name and the amount, the responsibility of the agent is compounded by the fact that in addition to understanding the unseen query, it has to reason about an unseen context pattern. In other words, the context also contains a form of turn that the agent has never seen during training, which is the first audit-more of the second example in Table~\ref{tab: One vs Multiple OOP}. For turn compression on the other hand, we did not observe a decrease. We can conjecture that memory networks are more resilient to turn compression even if they are present multiple times at the test time.


\begin{table}
 \centering
 \begin{tabular}{p{3cm}|p{1.5cm}|p{1.5cm}} \hline
 & \multicolumn{2}{|c}{Per-Response Accuracy}\\
 \hline
 Test Case & SMN & MMN\\
 \hline
 Turn Compression & 29.61 & 55.16\\
 Turn Compr. OOT & 29.07 & 55.13\\
 \hline
 Audit More & 10.50 & 15.02\\
 Audit More OOT & 10.90 & 15.43\\
 \hline
 \end{tabular}
 \caption{Multiple Out of Pattern per Dialog}
 \label{table : Multple OOP New}
\end{table}

\begin{table}[]
 \centering
 \begin{tabular}{p{0.95\columnwidth}}
 \hline
 \textbf{Sys} : I couldn't find the account you are looking for, would you like to change the account ?\\
 \textbf{User} : \textit{Yes. I want to use savings and change partner name to Michael. (The first occurrence of audit more)}\\
 \textbf{Sys} : There is no saved recipient with the name you provided, would you like to change the partner name ?\\
 \textbf{User} : \textit{Yes. Change it to Megan and change amount to 500 euros. (The second occurrence of audit more)}\\
 \hline
 
 \end{tabular}
 \caption{Example of Multiple OOP dialogue}
 \label{tab: One vs Multiple OOP}
\end{table}

\paragraph{The Easiest and the Hardest Challenge Cases} 

Finally, to investigate the difficulty of challenges that we have introduced with each OOP case, we should focus our attention to the easiest and the hardest cases. We observe that out of all the OOP test cases (both in-template and OOT settings) both memory networks performed quite poorly on handling new APIs. The results suggest that it is harder for the memory networks to interpret a new combination of slots and issue the related API calls. 
This could be partially explained by the position of the new API cases in the dialogue.
By design, new API cases happen at the beginning of the conversation (i.e. giving an intent together with unexpected slots, see example in Table \ref{tab:New API}). Therefore, the system has no context (no interaction memory) to reason on while selecting the response. On the contrary, for the easier turn compression case, the memory network is already expecting a possible change in the slot value (e.g. ``\textit{do you want to change the amount?"}) in the following turns, regardless of receiving it in the respective turn or in the next few. In fact, the network is already `primed' on selecting an amount related API call. Consequently, the memory networks have a better performance on turn compression rather than new API challenge. 

\section{Conclusions and Future Work}

In this paper, we explored some challenges connected to dataset creation for conversational agents and interpretability of neural models. 
In particular, we propose a methodology to create 
rich datasets for training end-to-end conversational agents and challenge them on unseen patterns at test time. We then experimented with Memory Networks and investigated their performance on the custom test cases. The apparently low accuracy levels on unseen challenge cases suggest that the synthetic data and challenge generation for low resource dialogue domains can act as a reasonable approximate to real life challenges in such domains. In other words, the more a dialogue model is able to handle these diverse challenges, the more it will be able to handle the unstructured or less structured dialogues in real human-machine interaction. As a future work we would like to test further neural models and create additional OOP challenge sets, even combining additional configurations. 

\bibliography{acl2017}

\begin{thebibliography}{}
\expandafter\ifx\csname natexlab\endcsname\relax\def\natexlab#1{#1}\fi

\bibitem[{Bai et~al.(2009)Bai, Weston, Grangier, Collobert, Sadamasa, Qi,
  Chapelle, and Weinberger}]{bai2009supervised}
Bing Bai, Jason Weston, David Grangier, Ronan Collobert, Kunihiko Sadamasa,
  Yanjun Qi, Olivier Chapelle, and Kilian Weinberger. 2009.
\newblock Supervised semantic indexing.
\newblock In {\em Proceedings of the 18th ACM conference on Information and
  knowledge management\/}. ACM, pages 187--196.

\bibitem[{Belinkov and Glass(To appear)}]{belinkov:2018:tacl}
Yonatan Belinkov and James Glass. To appear.
\newblock Analysis methods in neural language processing: A survey.
\newblock {\em Transactions of the Association for Computational Linguistics
  (TACL)\/} .

\bibitem[{Bordes et~al.(2017)Bordes, Boureau, and Weston}]{bordes2016learning}
Antoine Bordes, Y{-}Lan Boureau, and Jason Weston. 2017.
\newblock \href{https://openreview.net/forum?id=S1Bb3D5gg}{Learning end-to-end
  goal-oriented dialog}.
\newblock In {\em 5th International Conference on Learning Representations,
  {ICLR} 2017, Toulon, France, April 24-26, 2017, Conference Track
  Proceedings\/}.
\newblock
  \href{https://openreview.net/forum?id=S1Bb3D5gg}{https://openreview.net/forum?id=S1Bb3D5gg}.

\bibitem[{Cho et~al.(2014)Cho, Van~Merri{\"e}nboer, Bahdanau, and
  Bengio}]{cho2014properties}
Kyunghyun Cho, Bart Van~Merri{\"e}nboer, Dzmitry Bahdanau, and Yoshua Bengio.
  2014.
\newblock On the properties of neural machine translation: Encoder-decoder
  approaches.
\newblock {\em arXiv preprint arXiv:1409.1259\/} .

\bibitem[{Cooper et~al.(1996)Cooper, Crouch, Van~Eijck, Fox, Van~Genabith,
  Jaspars, Kamp, Milward, Pinkal, Poesio et~al.}]{cooper1996using}
Robin Cooper, Dick Crouch, Jan Van~Eijck, Chris Fox, Johan Van~Genabith, Jan
  Jaspars, Hans Kamp, David Milward, Manfred Pinkal, Massimo Poesio, et~al.
  1996.
\newblock Using the framework.
\newblock Technical report, Technical Report LRE 62-051 D-16, The FraCaS
  Consortium.

\bibitem[{Danescu-Niculescu-Mizil and Lee(2011)}]{danescu2011chameleons}
Cristian Danescu-Niculescu-Mizil and Lillian Lee. 2011.
\newblock Chameleons in imagined conversations: A new approach to understanding
  coordination of linguistic style in dialogs.
\newblock In {\em Proceedings of the 2nd Workshop on Cognitive Modeling and
  Computational Linguistics\/}. Association for Computational Linguistics,
  pages 76--87.

\bibitem[{Dodge et~al.(2016)Dodge, Gane, Zhang, Bordes, Chopra, Miller, Szlam,
  and Weston}]{dodge2016}
Jesse Dodge, Andreea Gane, Xiang Zhang, Antoine Bordes, Sumit Chopra,
  Alexander~H. Miller, Arthur Szlam, and Jason Weston. 2016.
\newblock \href{http://arxiv.org/abs/1511.06931}{Evaluating prerequisite
  qualities for learning end-to-end dialog systems}.
\newblock In {\em 4th International Conference on Learning Representations,
  {ICLR} 2016, San Juan, Puerto Rico, May 2-4, 2016, Conference Track
  Proceedings\/}.
\newblock
  \href{http://arxiv.org/abs/1511.06931}{http://arxiv.org/abs/1511.06931}.

\bibitem[{Eric and Manning(2017)}]{eric2017key}
Mihail Eric and Christopher~D Manning. 2017.
\newblock Key-value retrieval networks for task-oriented dialogue.
\newblock {\em arXiv preprint arXiv:1705.05414\/} .

\bibitem[{Guerini et~al.(2018)Guerini, Magnolini, Balaraman, and
  Magnini}]{guerini2018toward}
Marco Guerini, Simone Magnolini, Vevake Balaraman, and Bernardo Magnini. 2018.
\newblock Toward zero-shot entity recognition in task-oriented conversational
  agents.
\newblock In {\em Proceedings of the 19th Annual SIGdial Meeting on Discourse
  and Dialogue\/}. pages 317--326.

\bibitem[{Henderson et~al.(2014)Henderson, Thomson, and
  Williams}]{henderson2014second}
Matthew Henderson, Blaise Thomson, and Jason~D Williams. 2014.
\newblock The second dialog state tracking challenge.
\newblock In {\em Proceedings of the 15th Annual Meeting of the Special
  Interest Group on Discourse and Dialogue (SIGDIAL)\/}. pages 263--272.

\bibitem[{Hochreiter and Schmidhuber(1997)}]{hochreiter1997long}
Sepp Hochreiter and J{\"u}rgen Schmidhuber. 1997.
\newblock Long short-term memory.
\newblock {\em Neural computation\/} 9(8):1735--1780.

\bibitem[{Hopcroft et~al.(2001)Hopcroft, Motwani, and
  Ullman}]{hopcroft2001introduction}
John~E Hopcroft, Rajeev Motwani, and Jeffrey~D Ullman. 2001.
\newblock Introduction to automata theory, languages, and computation.
\newblock {\em Acm Sigact News\/} 32(1):60--65.

\bibitem[{Isbell et~al.(2000)Isbell, Kearns, Kormann, Singh, and
  Stone}]{isbell2000cobot}
Charles~Lee Isbell, Michael Kearns, Dave Kormann, Satinder Singh, and Peter
  Stone. 2000.
\newblock Cobot in lambdamoo: A social statistics agent.
\newblock In {\em AAAI/IAAI\/}. pages 36--41.

\bibitem[{Jafarpour et~al.(2010)Jafarpour, Burges, and
  Ritter}]{jafarpour2010filter}
Sina Jafarpour, Christopher~JC Burges, and Alan Ritter. 2010.
\newblock Filter, rank, and transfer the knowledge: Learning to chat.
\newblock {\em Advances in Ranking\/} 10:2329--9290.

\bibitem[{Jur{\v{c}}{\'\i}{\v{c}}ek et~al.(2011)Jur{\v{c}}{\'\i}{\v{c}}ek,
  Keizer, Ga{\v{s}}i{\'c}, Mairesse, Thomson, Yu, and
  Young}]{jurvcivcek2011real}
Filip Jur{\v{c}}{\'\i}{\v{c}}ek, Simon Keizer, Milica Ga{\v{s}}i{\'c}, Francois
  Mairesse, Blaise Thomson, Kai Yu, and Steve Young. 2011.
\newblock Real user evaluation of spoken dialogue systems using amazon
  mechanical turk.
\newblock In {\em Twelfth Annual Conference of the International Speech
  Communication Association\/}.

\bibitem[{Kelley(1984)}]{kelley1984iterative}
John~F Kelley. 1984.
\newblock An iterative design methodology for user-friendly natural language
  office information applications.
\newblock {\em ACM Transactions on Information Systems (TOIS)\/} 2(1):26--41.

\bibitem[{King and Falkedal(1990)}]{king1990using}
Margaret King and Kirsten Falkedal. 1990.
\newblock Using test suites in evaluation of machine translation systems.
\newblock In {\em COLNG 1990 Volume 2: Papers presented to the 13th
  International Conference on Computational Linguistics\/}. volume~2.

\bibitem[{Lehmann et~al.(1996)Lehmann, Oepen, Regnier-Prost, Netter, Lux,
  Klein, Falkedal, Fouvry, Estival, Dauphin et~al.}]{lehmann1996tsnlp}
Sabine Lehmann, Stephan Oepen, Sylvie Regnier-Prost, Klaus Netter, Veronika
  Lux, Judith Klein, Kirsten Falkedal, Frederik Fouvry, Dominique Estival, Eva
  Dauphin, et~al. 1996.
\newblock Tsnlp: Test suites for natural language processing.
\newblock In {\em Proceedings of the 16th conference on Computational
  linguistics-Volume 2\/}. Association for Computational Linguistics, pages
  711--716.

\bibitem[{Lowe et~al.(2015)Lowe, Pow, Serban, and Pineau}]{lowe2015ubuntu}
Ryan Lowe, Nissan Pow, Iulian Serban, and Joelle Pineau. 2015.
\newblock The ubuntu dialogue corpus: A large dataset for research in
  unstructured multi-turn dialogue systems.
\newblock {\em arXiv preprint arXiv:1506.08909\/} .

\bibitem[{Marietto et~al.(2013)Marietto, de~Aguiar, Barbosa, Botelho, Pimentel,
  Fran{\c{c}}a, and da~Silva}]{marietto2013artificial}
Maria das Gra{\c{c}}as~Bruno Marietto, Rafael~Varago de~Aguiar, Gislene
  de~Oliveira Barbosa, Wagner~Tanaka Botelho, Edson Pimentel, Robson dos~Santos
  Fran{\c{c}}a, and Vera~L{\'u}cia da~Silva. 2013.
\newblock Artificial intelligence markup language: A brief tutorial.
\newblock {\em arXiv preprint arXiv:1307.3091\/} .

\bibitem[{Raux et~al.(2005)Raux, Langner, Bohus, Black, and
  Eskenazi}]{raux2005let}
Antoine Raux, Brian Langner, Dan Bohus, Alan~W Black, and Maxine Eskenazi.
  2005.
\newblock Let's go public! taking a spoken dialog system to the real world.
\newblock In {\em Ninth European conference on speech communication and
  technology\/}.

\bibitem[{Ritter et~al.(2011)Ritter, Cherry, and Dolan}]{ritter2011data}
Alan Ritter, Colin Cherry, and William~B Dolan. 2011.
\newblock Data-driven response generation in social media.
\newblock In {\em Proceedings of the conference on empirical methods in natural
  language processing\/}. Association for Computational Linguistics, pages
  583--593.

\bibitem[{Serban et~al.(2016)Serban, Sordoni, Bengio, Courville, and
  Pineau}]{serban2016building}
Iulian~Vlad Serban, Alessandro Sordoni, Yoshua Bengio, Aaron~C Courville, and
  Joelle Pineau. 2016.
\newblock Building end-to-end dialogue systems using generative hierarchical
  neural network models.
\newblock In {\em AAAI\/}. volume~16, pages 3776--3784.

\bibitem[{Serban et~al.(2017)Serban, Sordoni, Lowe, Charlin, Pineau, Courville,
  and Bengio}]{serban2017hierarchical}
Iulian~Vlad Serban, Alessandro Sordoni, Ryan Lowe, Laurent Charlin, Joelle
  Pineau, Aaron~C Courville, and Yoshua Bengio. 2017.
\newblock A hierarchical latent variable encoder-decoder model for generating
  dialogues.
\newblock In {\em AAAI\/}. pages 3295--3301.

\bibitem[{Shah et~al.(2018)Shah, Hakkani-T{\"u}r, T{\"u}r, Rastogi, Bapna,
  Nayak, and Heck}]{shah2018building}
Pararth Shah, Dilek Hakkani-T{\"u}r, Gokhan T{\"u}r, Abhinav Rastogi, Ankur
  Bapna, Neha Nayak, and Larry Heck. 2018.
\newblock Building a conversational agent overnight with dialogue self-play.
\newblock {\em arXiv preprint arXiv:1801.04871\/} .

\bibitem[{Sordoni et~al.(2015)Sordoni, Galley, Auli, Brockett, Ji, Mitchell,
  Nie, Gao, and Dolan}]{sordoni2015neural}
Alessandro Sordoni, Michel Galley, Michael Auli, Chris Brockett, Yangfeng Ji,
  Margaret Mitchell, Jian-Yun Nie, Jianfeng Gao, and Bill Dolan. 2015.
\newblock A neural network approach to context-sensitive generation of
  conversational responses.
\newblock {\em arXiv preprint arXiv:1506.06714\/} .

\bibitem[{Sukhbaatar et~al.(2015)Sukhbaatar, Weston, Fergus
  et~al.}]{sukhbaatar2015end}
Sainbayar Sukhbaatar, Jason Weston, Rob Fergus, et~al. 2015.
\newblock End-to-end memory networks.
\newblock In {\em Advances in neural information processing systems\/}. pages
  2440--2448.

\bibitem[{Vinyals and Le(2015)}]{vinyals2015neural}
Oriol Vinyals and Quoc Le. 2015.
\newblock A neural conversational model.
\newblock {\em arXiv preprint arXiv:1506.05869\/} .

\bibitem[{Wang et~al.(2018)Wang, Singh, Michael, Hill, Levy, and
  Bowman}]{wang2018glue}
Alex Wang, Amapreet Singh, Julian Michael, Felix Hill, Omer Levy, and Samuel~R
  Bowman. 2018.
\newblock Glue: A multi-task benchmark and analysis platform for natural
  language understanding.
\newblock {\em arXiv preprint arXiv:1804.07461\/} .

\bibitem[{Weston et~al.(2014)Weston, Chopra, and Bordes}]{Weston2014MemoryN}
Jason Weston, Sumit Chopra, and Antoine Bordes. 2014.
\newblock Memory networks.
\newblock {\em CoRR\/} abs/1410.3916.

\bibitem[{Yu et~al.(2015)Yu, Bohus, and Horvitz}]{yu2015incremental}
Zhou Yu, Dan Bohus, and Eric Horvitz. 2015.
\newblock Incremental coordination: Attention-centric speech production in a
  physically situated conversational agent.
\newblock In {\em Proceedings of the 16th Annual Meeting of the Special
  Interest Group on Discourse and Dialogue\/}. pages 402--406.

\bibitem[{Zhao and Eskenazi(2018)}]{zhao2018zero}
Tiancheng Zhao and Maxine Eskenazi. 2018.
\newblock Zero-shot dialog generation with cross-domain latent actions.
\newblock {\em arXiv preprint arXiv:1805.04803\/} .

\bibitem[{Zue et~al.(1994)Zue, Seneff, Polifroni, Phillips, Pao, Goodine,
  Goddeau, and Glass}]{zue1994pegasus}
Victor Zue, Stephanie Seneff, Joseph Polifroni, Michael Phillips, Christine
  Pao, David Goodine, David Goddeau, and James Glass. 1994.
\newblock Pegasus: A spoken dialogue interface for on-line air travel planning.
\newblock {\em Speech Communication\/} 15(3-4):331--340.

\end{thebibliography}
\bibliographystyle{acl_natbib}

\end{document}